\documentclass{article}

\usepackage{microtype}
\usepackage{graphicx}
\usepackage{subfigure}
\usepackage{booktabs} 

\usepackage{hyperref}

\usepackage[accepted]{icml2020}

\icmltitlerunning{Prediction of CVDs from EHR using MT-GRU}

\usepackage{multirow}

\usepackage[acronym,toc]{glossaries}
\setkeys{glslink}{hyper=false}
\newacronym{ehr}{EHR}{electronic health record}
\newacronym{cvd}{CVD}{cardiovascular disease}
\newacronym{ml}{ML}{machine learning}
\newacronym{mt}{MT}{multi-task}
\newacronym{nhs}{NHS}{National Health Service}
\newacronym{lims}{LIMS}{Laboratory Information Management System}
\newacronym{bmi}{BMI}{body-mass index}
\newacronym{sbp}{SBP}{systolic blood pressure}
\newacronym{dbp}{DBP}{diastolic blood pressure}
\newacronym{rnn}{RNN}{recurrent neural network}
\newacronym{bnf}{BNF}{British National Formulary}
\newacronym{auc}{AUC}{area under the curve}
\newacronym{roc}{ROC}{receiver operating characteristic}
\newacronym{sen}{SEN}{sensitivity}
\newacronym{prec}{PRE}{precision}
\newacronym{alt}{ALT}{alanine aminotransferase}
\newacronym{mi}{MI}{myocardial infarction}

\newacronym{gru}{GRU}{gated recurrent unit}
\newacronym{mtgru}{MT-GRU}{multi-task GRU}
\newacronym{mtattgru}{MT-Att-GRU}{MT attention GRU}
\newacronym{lr}{LR}{logistic regression}
\newacronym{qrisk}{QRISK}{QRISK2}

\begin{document}

\twocolumn[
\icmltitle{Prediction of the onset of cardiovascular diseases \\ from electronic health records using multi-task gated recurrent units}

\icmlsetsymbol{equal}{*}

\begin{icmlauthorlist}
\icmlauthor{Fernando Andreotti}{equal,sh}
\icmlauthor{Frank S. Heldt}{sh}
\icmlauthor{Basel Abu-Jamous}{sh}
\icmlauthor{Ming Li}{sh}
\icmlauthor{Avelino Javer}{sh}
\icmlauthor{Oliver Carr}{sh}
\icmlauthor{Stojan Jovanovic}{sh}
\icmlauthor{Nadezda Lipunova}{sh}
\icmlauthor{Benjamin Irving}{sh}
\icmlauthor{Rabia T. Khan}{sh}
\icmlauthor{Robert D\"urichen}{equal,sh}
\end{icmlauthorlist}
\icmlaffiliation{sh}{Sensyne Health plc, Oxford, United Kingdom}
\icmlcorrespondingauthor{Rabia T. Khan}{rabia.khan@sensynehealth.com}

\icmlkeywords{RNN, Attention, EHR, Stroke, Myocardial Infarction}
\vskip 0.3in
]

\printAffiliationsAndNotice{\icmlEqualContribution} 

\begin{abstract}

In this work, we propose a \gls{mt} recurrent neural network with attention mechanism for predicting cardiovascular events from \glspl{ehr} at different time horizons. The proposed approach is compared to a standard clinical risk predictor (QRISK) and machine learning alternatives using 5-year data from a NHS Foundation Trust. The proposed model outperforms standard clinical risk scores in predicting stroke (AUC$=0.85$) and myocardial infarction (AUC$=0.89$), considering the largest time horizon. Benefit of using an \gls{mt} setting becomes visible for very short time horizons, which results in an AUC increase between $2-6\%$. Further, we explored the importance of individual features and attention weights in predicting \gls{cvd} events. Our results indicate that the recurrent neural network approach benefits from the hospital longitudinal information and demonstrates how \gls{ml} techniques can be applied to secondary care.

\end{abstract}

\section{Introduction}
\label{sec:int}

Cardiac events are associated with high mortality rates as well as long-lasting morbidities. They represent a significant burden to healthcare systems such as the UK's \gls{nhs} and continued efforts to improve risk prediction are of great importance as they allow preventive interventions \cite{Pike2016}. Existing clinical guidelines such as the QRISK2 \cite{Hippisley-Cox2008} score are often used in primary care settings for 10-year cardiovascular risk for an individual. These guidelines are  derived from populational studies and take into account well-established risk factors such as hypertension, cholesterol, age, smoking, and diabetes.

\gls{ehr} systems have become ubiquitous, with hospitals routinely collecting data at high frequencies and large numbers of variables \cite{Pike2016}. These systems contain longitudinal patient information like demographics, laboratory results, medications and diagnoses. Such rich information may provide a patient-tailored training set for predicting near-term risk of events when using state-of-the-art \gls{ml} approaches \cite{Goldstein2017}. However, \gls{ehr} analysis remains challenging due to its high dimensionality, noise, heterogeneity, sparsity and systematic biases. A large number of studies exist on \gls{ml}--based risk prediction approaches for patient deterioration in intensive care \cite{news2,johnson2013new} and population studies for primary care usage \cite{Hippisley-Cox2008}. 

Various approaches have been investigated to predict disease unspecific endpoints from \gls{ehr} such as length of stay, risk of re-admission or mortality. While many works attempt to predict future diagnoses from previous ones \cite{Lipton2015} or from clinical notes, few predict the risk of \glspl{cvd} based on \gls{ehr} data, such as stroke \cite{Teoh2018} or heart failure \cite{Choi2017}. In \citet{Teoh2018}, the author used \glspl{rnn} to predict the onset of stroke within a year of the last admission, reporting an \gls{auc} of $0.667$ by using temporal information of diagnoses codes and laboratory values of 8,000 patients. In a recent study, \citet{xu2019} applied an attention mechanism on top of a \gls{rnn} and evaluated their models in a multi-task learning setting for predicting the risk of stroke, \gls{mi} and death within a year. Adding attention resulted only in minor improvements, with the authors reporting an average increase of \gls{auc} by $0.065$. \citet{Choi2017} also proposed a \gls{rnn} with an attention mechanism to predict onset of heart failure. Common factors amongst these studies are i) use of limited number of features, e.g. only diagnoses and laboratory results, ii) the focus on a single prediction horizon, e.g.. the risk of having an event within a year after admission, and iii) no analysis of which features are most relevant for prediction.

In this work, we propose an algorithm to predict a patient's first diagnosis of ischemic stroke or \gls{mi}. For this purpose, a multi-task attention \gls{rnn} is proposed. The method takes as input the established risk factors as well as data of multiple \gls{ehr} modalities for prediction, including diagnoses, procedures, medications, encounter, demographic, laboratory and vital signs information. The main contributions of this paper are three-fold:
(i) Prediction of the onset of ischemic stroke or \gls{mi} for different time horizons in a single and multi-task setting; (ii) evaluation of the relevance of heterogeneous input data e.g. laboratory results and vital signs; and (iii) exploration of relevant observed events which lead to positive prediction using attention weights.

\section{Material and Methods}
\label{sec:materials}

Data was collected by the Oxford University Hospitals NHS Foundation Trust between 2014 -- 2019 as part of routine care. The longitudinal secondary care \gls{ehr} includes demographic information (i.e. sex, age), admission information (start/end dates, discharge method/destination, admission types - e.g.~in-patient, outpatient, emergency department), ICD-10 coded diagnoses, OPCS-4 coded procedures, medications as \gls{bnf} codes (prescribed both during visits and take-home), \gls{lims} (e.g.~blood and urine tests), digital bedside observations of vital signs (e.g.~systolic blood pressure, temperature, etc) as well as demographic measurements such as \gls{bmi}. 

In this study, we focus on two highly prevalent \glspl{cvd}, namely ischemic stroke (defined by ICD-10 codes I63.X or I69.3) and myocardial infarction (MI - code I21.X or I25.2). In particular, we include patients who have at least one observation at least 2 weeks prior to the \gls{cvd} event. Both cohorts are summarised in Table~\ref{tab:cohorts}. For each cohort, we construct a case-control group of patients who had never been diagnosed with the case \gls{cvd} (neither primary nor secondary reasons for admission). Controls were assigned by nearest neighbour matching on diseased patients on age, sex, and number of days containing observations. 

\begin{table}[t]
\label{sample-table}
\vskip 0.15in
\begin{center}
\begin{sc}
\scriptsize
  \caption{Patient numbers in each target cohort.}
  \label{tab:cohorts}
  \begin{tabular}{lcc}
    \toprule
    Time Horizon & Stroke cohort & MI cohort\\
    \hline
   1 month & 339 & 337 \\
   3 month & 910 & 951\\
   12 month & 1,727 & 1,959\\
   $>12$ month & 2,133 & 2,732 \\
    \bottomrule
\end{tabular}
\end{sc}
\end{center}
\end{table}

\subsection{Data representation}

Each patient is represented by a sequence of days in which observations are available, to which we refer as observation days. Data was aggregated into a single feature vector for each day. Continuous variables (e.g.~vital signs and laboratory results) are represented by the median, median absolute deviation, count of observations and an abnormality flag (indicating if values outside physiological ranges), whereas categorical variables (i.e. medication prescription) were counted and the 300 most common diagnoses and procedure codes were sparse encoded. Additional features representing the existence or absence of co-morbid conditions were incorporated by utilising the ICD-10 definition of the Charlson's co-morbidity score. Therefore, each patient is represented by a single 2D vector of $n_{days}$ and $n_{features}$, representing the number of observed days and features, respectively. Features with less than $1\,\%$ prevalence were excluded from our analysis. All features were scaled between $[0,1]$ and populational mean imputation was used on missing data points. 

\subsection{Patient trajectory prediction algorithms}
For the task of predicting \gls{cvd} events from a sequence of $n_{days}$ containing hospital observations, the following approaches are evaluated.

\subsubsection{Baseline clinical approach}

The \gls{qrisk} algorithm \cite{Hippisley-Cox2008} is the UK primary care standard clinical assessment tool, which calculates 10-year risk of cardiovascular disease. The score is based on Cox proportional hazards models and trained on well known populational risk factors. Since secondary care \gls{ehr} does not contain all risk factors from the \gls{qrisk} model, missing coefficients were set to zero. The method focuses on a single observation, therefore the last observation of each subject before the event was used.

\subsubsection{Baseline ML approach}

\Gls{lr} was used as baseline ML method using L1 regularisation to account for the sparse feature space. To allow a fair comparison and make use of the information contained on multiple past observed days, observations from previous timesteps were concatenated as additional features. We evaluated the inclusion of 1, 50 and 100 days, here denoted as LR-1, LR-50 and LR-100. 
\begin{table*}[t]
\centering
  \caption{Model performances for predicting myocardial infarction and ischemic stroke throughout all time horizons. Shown are average and standard deviation of \acrfull{auc}, \acrfull{sen}, and \acrfull{prec} of 5-fold cross validation. Best average results per time horizon and disease are highlighted in bold.}
  \label{tab:results}
  \tiny
  \begin{tabular}{|p{3.5em}l |c c c | c c c |}
    \toprule 
     Time  & Method & \multicolumn{3}{c}{Myocardial Infarction} & \multicolumn{3}{|c|}{Stroke} \\ \cline{3-8}
     horizon &   & \gls{auc}  & \acrshort{sen}  & \acrshort{prec} & \acrshort{auc}  & \acrshort{sen}  & \acrshort{prec} \\ 
    \hline 
\multirow{3}{*}{\shortstack[l]{1\\month}} &	\acrshort{mtgru}	&	\textbf{0.763	$\pm$	0.032}	&	0.694	$\pm$	0.124	&	0.138	$\pm$	0.016	&	
0.722	$\pm$	0.014	&	0.608	$\pm$	0.087	&	0.188	$\pm$	0.035	\\

&	\acrshort{mtattgru}	&	0.756	$\pm$	0.033	&	0.696	$\pm$	0.136	&	0.131	$\pm$	0.013	&	
\textbf{0.733	$\pm$	0.025}	&	0.684	$\pm$	0.101	&	0.165	$\pm$	0.024	\\

&	\acrshort{gru}	&	0.703	$\pm$	0.038	&	0.611	$\pm$	0.155	&	\textbf{0.727	$\pm$	0.082}	&	0.703	$\pm$	0.047	&	\textbf{0.694	$\pm$	0.144}	&	0.675	$\pm$	0.040	\\

&	LR-50	&	0.702	$\pm$	0.054	&	\textbf{0.772	$\pm$	0.136}	&	0.657	$\pm$	0.062	&	0.597	$\pm$	0.028	&	0.546	$\pm$	0.134	&	\textbf{0.631	$\pm$	0.020}	\\

&	\acrshort{qrisk} & 0.471	$\pm$	0.023	&	0.834	$\pm$	0.155	&	0.516	$\pm$	0.013	&	0.486	$\pm$	0.041	&	0.567	$\pm$	0.31	&	0.564	$\pm$	0.038	\\
\hline
\multirow{3}{*}{\shortstack[l]{3\\months}} & \acrshort{mtgru}	&	0.773	$\pm$	0.010	&	0.707	$\pm$	0.061	&	0.355	$\pm$	0.043	&
\textbf{0.734	$\pm$	0.025}	&	0.698	$\pm$	0.069	&	0.371	$\pm$	0.018	\\

&	\acrshort{mtattgru}	&	0.759	$\pm$	0.016	&	\textbf{0.734	$\pm$	0.100}	&	0.326	$\pm$	0.027	&	
0.732	$\pm$	0.025	&	0.683	$\pm$	0.091	&	0.363	$\pm$	0.017	\\

&	\acrshort{gru}	&	\textbf{0.788	$\pm$	0.008}	&	0.682	$\pm$	0.032	&	\textbf{0.769	$\pm$	0.012}	&	
0.726	$\pm$	0.037	&	0.654	$\pm$	0.122	&	\textbf{0.711	$\pm$	0.066}	\\

&	LR-50	&	0.740	$\pm$	0.024	&	0.678	$\pm$	0.094	&	0.693	$\pm$	0.020	&	
0.679	$\pm$	0.032	&	\textbf{0.699	$\pm$	0.092}	&	0.641	$\pm$	0.034	\\

&	\acrshort{qrisk}	&	0.474	$\pm$	0.037	&	0.681	$\pm$	0.205	&	0.516	$\pm$	0.013	&	0.479	$\pm$	0.035	&	0.431	$\pm$	0.326	&	0.556	$\pm$	0.058	\\
\hline
\multirow{3}{*}{\shortstack[l]{12\\months}}&	\acrshort{mtgru}	&	0.811	$\pm$	0.018	&	\textbf{0.794	$\pm$	0.045}	&	0.594	$\pm$	0.041	&	\textbf{0.801	$\pm$	0.010}	&	\textbf{0.765	$\pm$	0.065}	&	0.637	$\pm$	0.033	\\

&	\acrshort{mtattgru}	&	0.797	$\pm$	0.012	&	0.718	$\pm$	0.055	&	0.610	$\pm$	0.042	&	
0.780	$\pm$	0.013	&	0.647	$\pm$	0.033	&	0.672	$\pm$	0.015	\\

&	\acrshort{gru}	&	\textbf{0.850	$\pm$	0.011}	&	0.769	$\pm$	0.035	&	\textbf{0.789	$\pm$	0.029}	&	
0.793	$\pm$	0.024	&	0.724	$\pm$	0.089	&	\textbf{0.731	$\pm$	0.024}	\\

&	LR-50	&	0.758	$\pm$	0.005	&	0.660	$\pm$	0.034	&	0.714	$\pm$	0.014	&	
0.737	$\pm$	0.018	&	0.691	$\pm$	0.064	&	0.692	$\pm$	0.034	\\

&	\acrshort{qrisk}	&	0.475	$\pm$	0.02	&	0.72	$\pm$	0.241	&	0.509	$\pm$	0.004	&	0.472	$\pm$	0.025	&	0.466	$\pm$	0.357	&	0.574	$\pm$	0.072	\\
\hline
\multirow{3}{*}{\shortstack[l]{$>12$\\months}} &	\acrshort{mtgru}	&	\textbf{0.897	$\pm$	0.015}	&	\textbf{0.779	$\pm$	0.023}	&	
\textbf{0.847	$\pm$	0.027}	
&	0.849	$\pm$	0.009	&	0.774	$\pm$	0.030	&	0.768	$\pm$	0.018	\\

&	\acrshort{mtattgru}	&	0.885	$\pm$	0.014	&	0.769	$\pm$	0.028	&	0.824	$\pm$	0.029	&	
0.853	$\pm$	0.010	&	0.749	$\pm$	0.058	&	0.791	$\pm$	0.027	\\

&	\acrshort{gru}	&	0.891	$\pm$	0.009	&	0.777	$\pm$	0.013	&	0.845	$\pm$	0.008	&	
\textbf{0.868	$\pm$	0.004}	&	\textbf{0.768	$\pm$	0.053}	&	\textbf{0.799	$\pm$	0.039}	\\

&	LR-50	&	0.785	$\pm$	0.008	&	0.706	$\pm$	0.026	&	0.717	$\pm$	0.020	&	
0.785	$\pm$	0.017	&	0.686	$\pm$	0.055	&	0.745	$\pm$	0.017	\\

&	\acrshort{qrisk}	&	0.473	$\pm$	0.017	&	0.775	$\pm$	0.248	&	0.506	$\pm$	0.003	&	0.468	$\pm$	0.021	&	0.577	$\pm$	0.293	&	0.53	$\pm$	0.04	\\
    \bottomrule
  \end{tabular}
\end{table*}
\subsubsection{Proposed Recurrent Neural Network}

In order to leverage the longitudinal EHR data, we propose a \gls{mtgru} \gls{rnn}-based approach. In the single-task scenario, the model predicts \gls{cvd} in a single time horizon based on historical information provided. We further extend the model into a \gls{mtgru} prediction model such that, given the past history, the model predicts a sequence of $n_t$  time horizons simultaneously. Further, we use attention \cite{Luong2015} to assign weights to each day of a patient’s health record when predicting the target disease. The final model is referred to as \gls{mtattgru}. Hyperparameter search was performed using Bayesian optimisation. Zero padding was applied to pad sequences to $n_{days}$ which is also a hyperparameter of each model. Implementation details are available as Supplementary Material.

\subsubsection{Performance evaluation}
\label{sec:performance}

\gls{ml} and clinical approaches were assessed for 1, 3, 12 and $>12$ months time horizons using 5-fold cross-validation. Prediction accuracy is measured in terms of \gls{auc} of the \gls{roc} as well as \gls{sen}, and \gls{prec}. To further characterise the models, we calculated the permutation feature importance post-hoc as the change in  F$_1$ score after randomising each feature individually. In addition, we qualitatively assessed the attention weights of \gls{mtattgru} trained models to demonstrate how the models incorporate information about previous observation days.


\section{Results}

\begin{figure}[b!]
  \centering
  \subfigure[1 month\label{fig:roc_curves_short}]{\includegraphics[width=0.76\linewidth]{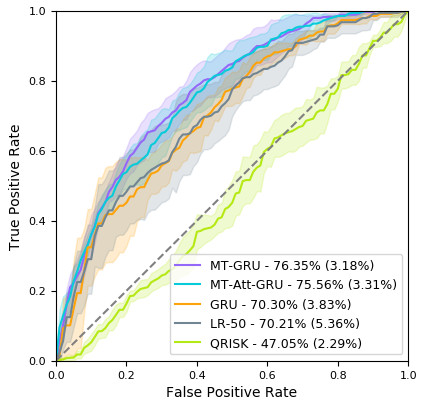}}\hfill
  
  \subfigure[$>12$ months]{\includegraphics[width=0.76\linewidth]{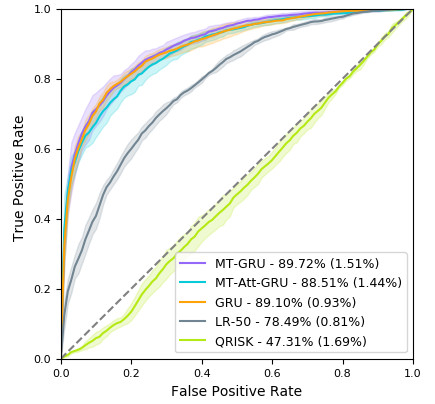}}\hfill
  
  \caption{Average \gls{roc} for the proposed models in comparison to \gls{lr} and \gls{qrisk} to predict the risk of having myocardial infarction. Predicted time horizons shown (a) 1 and (b) $>12$ months. Shaded areas represent the standard deviation based on 5-fold cross-validation.} 
  \label{fig:roc_curves}
\end{figure}

Table~\ref{tab:results} lists the \gls{auc}, \gls{sen} and \gls{prec} for all four investigated time horizons and both diseases. Due to space limitations, only the results for LR-50 are shown in Table \ref{tab:results}. Extended results are shown in Supplementary Materials. Overall, the proposed \gls{gru} models outperform \gls{lr} and \gls{qrisk} in all tasks. In shorter time horizons, in which data is limited (see Table~\ref{tab:cohorts}), both \gls{mtgru} and \gls{mtattgru} outperform the single-task \gls{gru}. In case of estimating the risk of having \gls{mi} (Figure~\ref{fig:roc_curves_short}) the mean AUC values for \gls{gru}, \gls{mtgru}, and \gls{mtattgru} are $0.703$, $0.763$, and $0.756$, respectively. Whereas in longer horizons \gls{gru} methods perform comparably. 

\begin{figure}[b!]
\vspace{-2em}
\centering
\subfigure[Permutation feature importance for \gls{mtgru}.\label{fig:feat_imp}]{\includegraphics[width=0.45\textwidth]{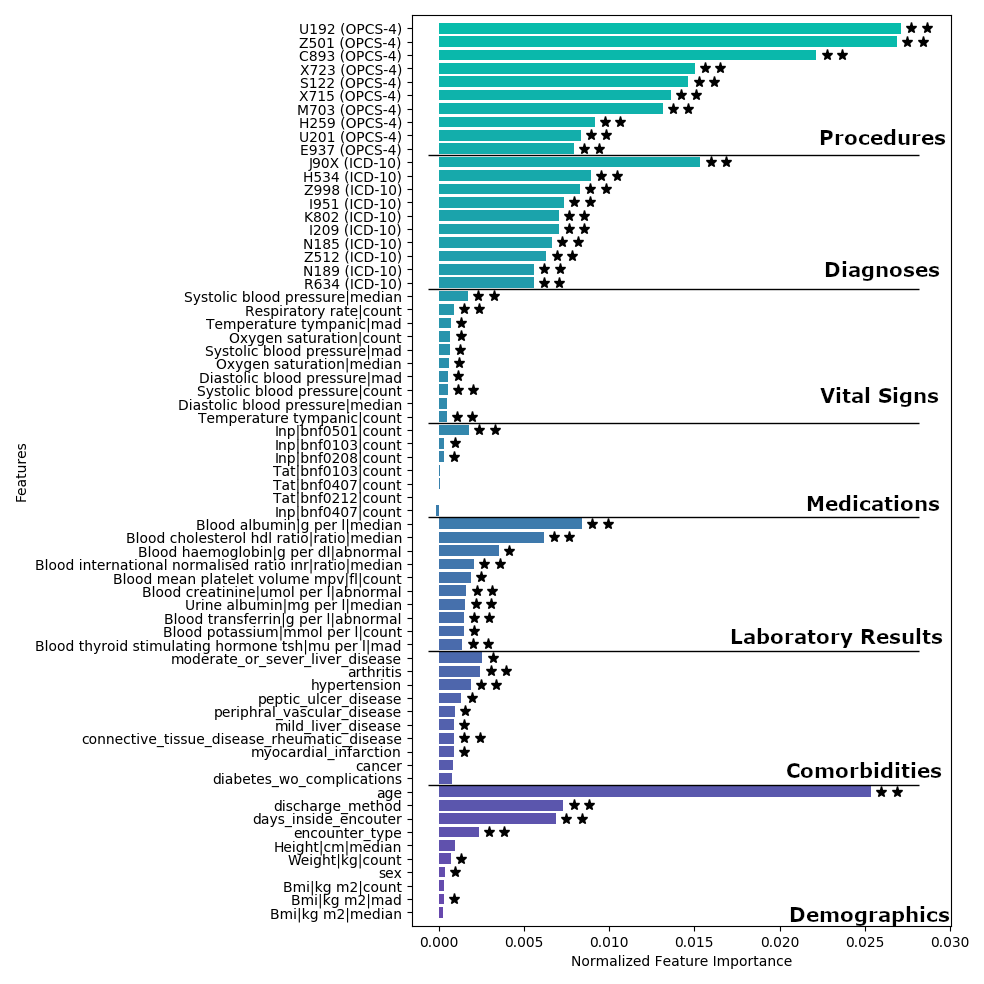}}

\subfigure[Attention weights in \gls{mtattgru}.\label{fig:attention}]{\includegraphics[width=0.45\textwidth, trim={0 3cm 0 7cm},clip]{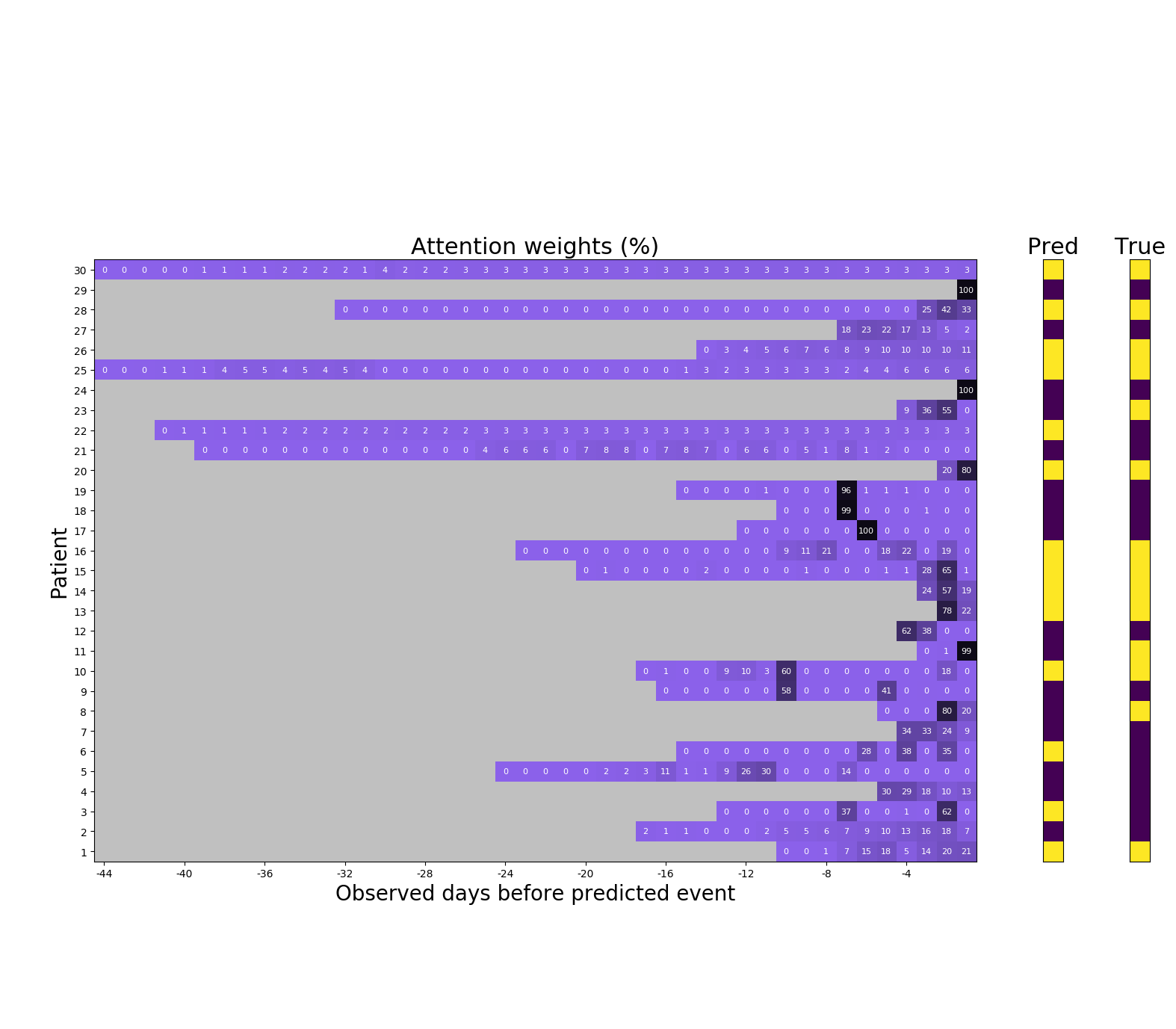}}\hfill
\caption{Feature importance (a) and attention weights (b) for \gls{mi} cohort with a $>12$ months time horizon. In (a) asterisks mark significance from zero with t-test p-value thresholds of 5\% ($\ast$) and 1\% ($\ast\ast$). At most, the top 10 features for each category is shown.} 
\label{fig:interpret}
\end{figure}

Next, we aim to further understand the model's decision. Figure~\ref{fig:feat_imp} shows the feature importance for the \gls{mtgru} model evaluated on the \gls{mi} cohort with a $>12$ months prediction horizon. Features such as age, previous albumin levels, diagnoses of pleural effusion (J90) and procedure codes for 24h Holter electrocardiogram (U19.2) are deemed important. Last, to explore the relevance of past \gls{ehr} information, we refer to the attention weights of \gls{mtattgru} in Fig.~\ref{fig:attention}. The model focuses on specific observation days throughout a patient's history. Overall, the model seems to weigh more recent observations higher.


\section{Discussion}

An important aspect of the proposed \gls{gru}-based approaches is that they are able to incorporate the longitudinal information present in \gls{ehr}. The proposed approaches outperform consistently clinical and \gls{ml} baselines. While multi-task approaches leverage the availability of data in shorter time horizons, in longer time frames they are not as helpful. The addition of attention does not improve model performance.
Meanwhile, the \gls{qrisk} \gls{auc} values are close to $0.5$ as expected due to the following reasons: i) the fact that some of the features are not available in \gls{ehr} and; ii) the algorithm is trained for population studies and should indeed under-perform in an age- and sex-matched setup.

Population models such as the \gls{qrisk} make use of features such as age, systolic blood pressure and \gls{bmi} as well-known indicators of future \gls{cvd}. Consequently, we expect models to use these variables. The feature importance analysis highlights potentially important predictors of future cardiac events. Reassuringly, the most important features broadly align with the current standard of care. Our results confirm the high relevance of diagnoses and procedure codes as used in previous studies \cite{xu2019}, but also indicate the benefit of integrating further \gls{ehr} modalities such as laboratory values and vital signs. This might be one reason why our GRU models achieve an AUC between $0.8-0.85$ for predicting the risk of \gls{mi} within a 12 month time horizon, in contrast to \cite{xu2019} who reported AUC values of $~0.78$. 


In addition to relevant features, attention is able to highlight past observation days which were most important in a model's decision as shown in Fig.~\ref{fig:attention}. It is reassuring to notice the \gls{mtattgru} does not only focus on recent observations, but also on historical observations. 

Although the results here presented are promising, our approach has several limitations, which we plan to address in the future. These limitations include:
\begin{itemize}
    \item \textit{Data representation:} Secondary care data is highly sparse. Hence, embeddings or graph networks may help to better represent intrinsic structures of \gls{ehr} data and improve predictions. 
    
    \item \textit{Multi-task learning:} Our results indicate that the \gls{mtgru} model can learn similarities across time horizons, which results in a superior prediction performance, in particular for scenarios with scarce training data. In the future, we plan to extend this approach for the task of predicting multiple diseases. 
    
    \item \textit{Multiple hospitals:} As most studies in the field \cite{xu2019, Teoh2018} our evaluation is restricted to a single NHS trust. To increase robustness of \gls{ml} models and assess generalisability, we plan to evaluate on further datasets and potentially investigate the usage of transfer learning approaches. 
\end{itemize}

\section{Conclusion}

This contribution presents a multi-task attention \gls{rnn} approach (\gls{mtattgru}) for predicting \gls{mi} and stroke events from the rich and longitudinal data available in \gls{ehr}. The method was evaluated using multivariate information from a \gls{nhs} trust. The proposed method outperforms baseline approaches and standard clinical tools for predicting \gls{cvd}s at different time horizons.

\section*{Acknowledgments}

This work uses data provided by patients and collected by the NHS as part of their care and support. We believe using patient data is vital to improve health and care for everyone and would, thus, like to thank all those involved for their contribution. The data were extracted, anonymised, and supplied by the Trust in accordance with internal information governance review, NHS Trust information governance approval, and General Data Protection Regulation (GDPR) procedures outlined under the Strategic Research Agreement (SRA) and relative Data Sharing Agreements (DSAs) signed by the Trust and Sensyne Health plc.

This research has been conducted using the Oxford University Hospitals NHS Foundation Trust Clinical Data Warehouse, which is supported by the NIHR Oxford Biomedical Research Centre and Oxford University Hospitals NHS Foundation Trust. Special thanks to Kerrie Woods, Kinga Varnai, Oliver Freeman, Hizni Salih, Zuzana Moysova, Professor Jim Davies and Steve Harris.

\bibliography{example_paper}
\bibliographystyle{icml2020}

\vfill
\clearpage

\end{document}